\documentclass[11pt]{article}

\usepackage{arxiv}

\usepackage[utf8]{inputenc} 
\usepackage[T1]{fontenc}    
\usepackage{hyperref}       
\usepackage{url}            
\usepackage{booktabs}       
\usepackage{amsfonts}       
\usepackage{nicefrac}       
\usepackage{microtype}      
\usepackage{graphicx}
\usepackage{natbib}
\usepackage{doi}
\usepackage{xcolor}         

\usepackage{amsmath}
\usepackage{amssymb}
\usepackage{float}
\usepackage{enumitem} 
\usepackage{subcaption}

\newif\ifblind
\blindfalse   

\newcommand{\blindtext}[2]{\ifblind #2\else #1\fi}

\title{Leveraging Convolutional and Graph Networks for an Unsupervised Remote Sensing Labelling Tool}
\date{January 27, 2026} 

\blindtext{
\author{\href{https://orcid.org/0009-0004-6699-9707}{Tulsi Patel\hspace{1mm}\raisebox{-0.2ex}{\includegraphics[scale=0.09]{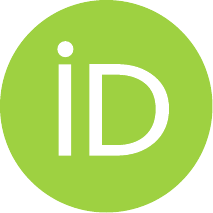}}} \\
	Department of Computer Science\\
	Swansea University\\
	Swansea \\
	\texttt{912100@Swansea.ac.uk} \\
	\And
	\href{https://orcid.org/0000-0001-8991-1190}{Mark W. Jones\hspace{1mm}\raisebox{-0.2ex}{\includegraphics[scale=0.09]{Images/orcid.pdf}}} \\
	Department of Computer Science\\
	Swansea University\\
	Swansea \\
	\texttt{M.W.Jones@Swansea.ac.uk} \\
	\And
	Thomas Redfern \\
	UK Hydrographic Office\\
	Taunton TA1 2DN, UK\\
	\texttt{Thomas.Redfern@UKHO.gov.uk} \\
}

}{
\author{Anonymous}

}

\begin{document}

\maketitle

\begin{abstract}
Machine learning for remote sensing imaging relies on up-to-date and accurate labels for model training and testing. Labelling remote sensing imagery is time and cost intensive, requiring expert analysis. Previous labelling tools rely on pre-labelled data for training in order to label new unseen data. In this work, we define an unsupervised pipeline for finding and labelling geographical areas of similar context and content within Sentinel-2 satellite imagery. Our approach removes limitations of previous methods by utilising segmentation with convolutional and graph neural networks to encode a more robust feature space for image comparison. Unlike previous approaches we segment the image into homogeneous regions of pixels that are grouped based on colour and spatial similarity. Graph neural networks are used to aggregate information about the surrounding segments enabling the feature representation to encode the local neighbourhood whilst preserving its own local information. This reduces outliers in the labelling tool, allows users to label at a granular level, and allows a rotationally invariant semantic relationship at the image level to be formed within the encoding space. 
Our pipeline achieves high contextual consistency, with similarity scores of SSIM = 0.96 and SAM = 0.21 under context-aware evaluation, demonstrating robust organisation of the feature space for interactive labelling.
\end{abstract}


\blindtext{}{\newpage}

\section{Introduction}

Rapid advancements in satellite imagery and Remote Sensing (RS) have led to an explosion of data, with the Sentinel-2 mission producing 1.6 terabytes of data daily. This wealth of data enables applications across agriculture, land use mapping, environmental monitoring, coastal management and monitoring and geological studies \citep{karthikeyan2020review,zhang2022artificial,weiss2020remote,ahmadi2021fault,SEALE2022113044} -- all of which require high quality labelled datasets.
However, the relatively large pixel resolutions of Sentinal-2 (covering 10 to 60m) create challenges for labelling. A single pixel, representing a 10m by 10m area, may cover both land and water, introducing ambiguity and mixed spectral responses. Traditional labelling approaches, whether bounding boxes, pixel-level masks, or small image segments, are time consuming and costly, requiring expert input.

\subsection*{Existing Labelling Tools}
Several tools have been developed to address these challenges. For example, \citeauthor{li2021labelrs} introduced LabelRS \citep{li2021labelrs}, which leverages OpenStreetMaps, Land Use Land Cover (LULC) datasets, and ArcGIS annotations to generate semantic, object, or image classification datasets. While effective, such tools rely on predefined label categories, limiting flexibility and adaptability to new or unseen classes. Beyond LabelRS, most existing solutions are not full labelling platforms but retrieval-based systems that assist annotation by retrieving and presenting similar images to a target image. These include CBIR-driven frameworks \citep{qi2020mlrsnet,hou2019v} and deep learning extensions such as NetVLAD and SatResNet-50 \citep{napoletano2018visual}, which improve retrieval accuracy but still depend on prior labels for supervised training. More recently, \citeauthor{patel2023manifold} \citep{patel2023manifold} proposed an interactive labelling tool based on manifold embeddings (UMAP and t-SNE), enabling experts to explore and annotate clusters in a 2D space. However, this approach struggles with outliers and lacks contextual awareness, highlighting the need for more robust, context-sensitive labelling solutions.

\subsection*{Methods Used in These Tools}
To improve labelling efficiency, many approaches borrow from \textbf{Content-Based Image Retrieval (CBIR)}, which retrieves similar content through three main steps: feature representation, indexing, and similarity measurement \citep{LI202194}. CBIR-based systems aim to present visually similar images to assist annotation, assuming that retrieved content shares sufficient semantic similarity.

\paragraph{Traditional Feature Extraction.}
Early CBIR methods relied on low-level descriptors capturing texture, colour, and shape characteristics. Common techniques include Scale-Invariant Feature Transform (SIFT) \citep{chen2020iterative,zhu2016bag}, Gray-Level Co-occurrence Matrices (GLCM), wavelets, Gabor filters, and Local Binary Patterns (LBP) \citep{huang2014multichannel,ranchin1993wavelet,yang2008comparing,huang2016remote}. These techniques primarily focus on either texture features (which capture spectral relationships) or the spectra themselves. Once extracted, these descriptors are often aggregated into mid-level representations using algorithms such as Bag of Visual Words (BoW) and Vector of Locally Aggregated Descriptors (VLAD) \citep{jegou2010aggregating,sivic2003video,cao2010spatial}. BoW employs K-means clustering to create a visual codebook and histograms of local features, while VLAD improves upon BoW by encoding spatial relationships between features and cluster centres. Despite their popularity, these methods typically lose semantic relationships between image features, limiting their effectiveness for complex remote sensing imagery.

\paragraph{Deep Learning-Based Retrieval.}
Deep Convolutional Neural Networks (DCNNs) have transformed image retrieval by learning hierarchical features that outperform handcrafted descriptors \citep{song2019survey,rs11050493}. Architectures such as VGG, ResNet, OverFeat, and CaffeNet have been widely adopted \citep{wang2017aggregating}, with extensions like NetVLAD and SatResNet-50 achieving state-of-the-art performance in remote sensing retrieval tasks \citep{napoletano2018visual}. These models aggregate convolutional features into compact descriptors, often combined with VLAD encoding for improved spatial awareness \citep{wang2017aggregating}. Unsupervised variants using autoencoders, generative models, and hybrid approaches with SVMs or visual codebooks further reduce reliance on labelled data \citep{romero2015unsupervised,rs10081243,doi:10.1080/2150704X.2020.1731769}. However, even deep models can struggle with contextual relationships when features are treated independently.

\paragraph{Graph-Based and Context-Aware Methods.}
To address the semantic gap, recent work incorporates Graph Neural Networks (GNNs) to encode spatial context and multi-label co-occurrence relationships. For example, \citeauthor{sumbul2021novel} \citep{sumbul2021novel} combine CNN feature extraction with GNN-based similarity refinement using multiple Siamese networks and triplet loss. \citeauthor{WANG2021107785} \citep{WANG2021107785} propose a deep hashing approach using Graph Neural Networks (GNNs), where deep features are extracted from a pre-trained ResNet network. Other studies leverage GNNs for cross-modal matching (e.g., text-image retrieval) and multi-label image retrieval, demonstrating improvements when attention mechanisms are introduced \citep{yu2022text,9173783}. Contrastive learning and triplet loss have also been applied to enforce neighbourhood consistency in embedding spaces \citep{9353191,yan2020image,draganov2023unexplainable}. These approaches highlight the importance of preserving local context, such as adjacency to water or urban density, when organising feature spaces for retrieval and labelling. \citeauthor{8089668} identified a gap between low-level features and high-level semantic concepts \citep{8089668}. They proposed a sub-graph matching strategy that emphasizes neighbourhood contexts within segmented images. While effective, this approach is computationally expensive \citep{8089668}. More recent work \citep{9526616} improves on this by using contrastive loss between Siamese Graph Convolutional Networks (GCNs), replacing sub-graph matching. This method applies context-based attention to nodes, enhancing image retrieval performance. The key advantage of GCNs is their ability to encode neighbourhood information into each segment through message-passing operations, facilitating more context-aware feature extraction.

\paragraph{Manifold Embedding and Interactive Labelling.}
Dimensionality reduction techniques such as t-SNE and UMAP have been explored to visualise high-dimensional feature spaces for interactive labelling. \citeauthor{halladin2019t} \citep{halladin2019t} applied t-SNE to enhance reference data for heterogeneous vegetation. Their method provides contextual feedback about the relationships between features, such as distinguishing urban areas from surrounding land or dense urban zones. This contextual discrimination, based on geographical proximity, allows labellers to customise datasets more effectively, ensuring they can balance and fine-tune their labelled datasets according to specific requirements. \citeauthor{patel2023manifold} \citep{patel2023manifold} proposed an interactive tool that embeds features into a 2D space for exploration and annotation. These techniques are useful for preserving local distances between data points when reducing dimensionality, which aids in visualising machine learning results \citep{TimeCluster2019,ConcurrentSelections}. Their approach enables experts to label clusters visually and interactively but suffers from outliers and orientation sensitivity, where strong texture gradients (e.g., land-water boundaries) distort clustering. These limitations underscore the need for embeddings that are both context-aware and rotationally invariant.

Recent work by \citeauthor{SUN2025241} \citep{SUN2025241} and \citeauthor{SUN2025241b} \citep{SUN2025241b} introduces highly efficient frameworks for unsupervised change detection in heterogeneous and multimodal remote sensing imagery. RIEM \citep{SUN2025241} leverages rule-based constraints within superpixels to detect changes without explicit image comparison, while LPEM\citep{SUN2025241b} preserves locality and spatial continuity through an energy-based formulation, avoiding intermediate difference images. These are highly efficient and improve robustness to imaging conditions by avoiding complex transformations for aligning heterogeneous data. Both methods share our goal of reducing reliance on predefined labels and exploiting spatial context. In contrast, our approach targets unsupervised labelling and clustering rather than change detection, learning context-aware embeddings via CNN and GNN to enable interactive exploration and semantic organisation of large datasets.

\subsection*{Limitations of Existing Approaches}
Overall, these studies demonstrate a progression from low-level descriptors to deep learning and graph-based approaches, with increasing emphasis on contextual relationships and interactive labelling. However, existing methods still face challenges in semantic organisation, robustness to outliers, and adaptability to new classes -- motivating our proposed pipeline that integrates CNNs and GNNs for context-aware, unsupervised labelling.

Existing methods face key limitations: (i) dependence on predefined label categories, reducing adaptability; (ii) loss of semantic relationships in codebook-based approaches; (iii) sensitivity to outliers and orientation, which can distort clustering; and (iv) insufficient encoding of spatial context, which is critical for remote sensing semantics. These gaps motivate the need for a more robust, context-aware solution that supports unsupervised labelling and interactive exploration.

\subsection*{Our Contribution}
To address the challenge of costly and time-consuming labelling in remote sensing, we sought to develop a toolchain that enhances efficiency by presenting experts with visually and contextually similar content for rapid annotation. Inspired by Content-Based Image Retrieval (CBIR), which retrieves similar images through feature representation, indexing, and similarity measurement \citep{LI202194}, we extend this concept to an unsupervised setting where semantic organisation is achieved without predefined labels. Unlike prior approaches that rely on low-level descriptors or purely deep learning-based embeddings, our method integrates Convolutional Neural Networks (CNNs) for rich feature extraction and Graph Neural Networks (GNNs) for encoding neighbourhood relationships, enabling context-aware embeddings that preserve spatial semantics. This design mitigates the impact of outliers, introduces rotational invariance, and supports interactive exploration of large datasets. In contrast to existing manifold embedding tools \citep{patel2023manifold}, which suffer from feature misalignment and orientation sensitivity, our pipeline provides a more robust and reliable foundation for visual labelling.

Specifically, we make the following contributions:
\begin{itemize}
    \item[\textbf{A)}] We improve the coherence of embedding spaces by achieving tighter cluster formation, facilitating easier navigation through unsupervised clustering.
    \item[\textbf{B)}] By integrating GNNs and feature encoding on image segmentations, we demonstrate that the resulting two-dimensional embedding space is robust to spatial transformations and encodes local context.
    \item[\textbf{C)}] Finally, we apply this methodology to develop an advanced interactive tool for exploration and labelling, enhancing both efficiency and accuracy in dataset creation.
\end{itemize}

\section{Materials and Methods}

\begin{figure*}[t]
  \centering
   \includegraphics[width= 1.0 \textwidth ]{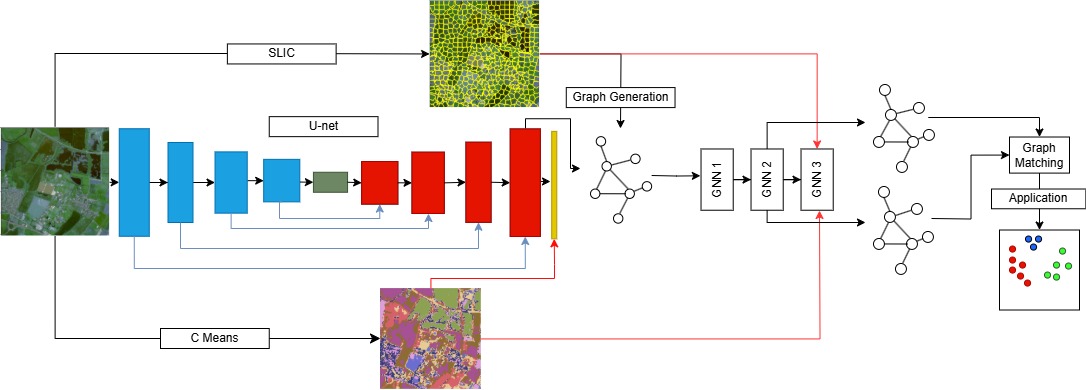}
   \caption{This diagram illustrates the data flow within the pipeline. Blue boxes represent the contracting layers of the U-Net, red boxes the expansive layers, and yellow the predictive layer. Blue lines indicate skip connections. Red lines show how data is used in the loss function, where the third GNN layer reshapes its input to enable comparison with the targets.}
   \label{fig:pipeline draft}
\end{figure*}

The proposed pipeline (Figure~\ref{fig:pipeline draft}) consists of three key components: a U-Net for feature extraction, a Graph Neural Network (GNN) for generating relational embeddings, and a graph-matching algorithm for assessing graph similarity. This design explicitly bridges the gap between low-level pixel features and high-level semantic concepts as identified by \citeauthor{8089668} \citep{8089668}. First, SLIC segmentation operates directly on raw pixel values to partition the image into super-pixels, creating spatially coherent regions that serve as graph nodes. These segments preserve local spectral and textural characteristics while introducing a structural representation of the image. Next, the U-Net extracts rich activation maps from the original image, which are aggregated within each segment to form node-level feature vectors. The GNN then propagates information across these nodes, encoding neighbourhood relationships such as adjacency to water or urban density. Through this message-passing process, isolated low-level descriptors are transformed into context-aware embeddings that capture both local detail and spatial semantics. Finally, graph matching and UMAP projection organise these enriched embeddings into a manifold where clusters correspond to meaningful categories (e.g., vegetation, urban, coastal), enabling interactive labelling at both chip and segment levels. This progression from raw pixels to graph-based context ensures that local pixel information is contextualised within its environment, achieving the bridging between low-level and high-level representations.

\subsection{Dataset}

We utilise the Sentinel-2 Water Edges Dataset (SWED) developed by the UK Hydrographic Office \citep{SEALE2022113044}. It consists of globally distributed Sentinel-2 scenes captured 2017--2021
annotated for coastline extraction, although we do not use the labels. The images are tiles, regions of 10,980m by 10,980m, taken in a single pass of the satellite. Minimal cloud coverage is present within the dataset, and each tile contains a diverse set of coastal landscape features from globally distributed locations.

\subsection{Fuzzy C-Means target extraction}
Due to the spatial resolution of most satellite platforms, pixel assignment can be ambiguous or uncertain \citep{zhang2024fast}. Fuzzy C-means extraction, unlike its K-means counterpart, assigns to each data point a degree of membership to each cluster, transforming the problem into a multi-label classification. We use all 12 spectral channels for clustering. To reduce memory requirements during model training, we apply uniform subsampling by selecting every $56^{\text{th}}$ pixel across the image, ensuring representative coverage while significantly lowering computational cost. After training, the clustering model is applied to the full-resolution image to obtain tile-wide classifications. The resultant centroids do not correspond to specific physical materials (e.g., water or concrete) but rather represent dominant spectral patterns, combinations of reflectance values across bands, that capture the most common pixel properties in the scene. These properties reflect typical spectral signatures rather than discrete land-cover classes, enabling soft classification and generalisation for pixels that do not fully belong to any single cluster.

\subsection{Pre-Proccessing}
For processing by the U-Net and GNN we split the tile into chips. A chip is created by applying a sliding window of size $256 \times 256$ pixels to the larger image, with no overlap between consecutive windows. This results in $42\times 42$ chips containing $10,752 \times 10,752$ of the original tiles $10,980 \times 10,980$ pixels. For training and testing, we split the dataset by taking every fourth chip in a top-down, row-by-row pattern for the test set obtaining a 75/25 split. The size of the chip is a balance between training size and preserving contextual neighbourhood information. The larger the image the more computationally expensive the resulting pipeline becomes however more information about neighbouring features can be encoded. Considering CNN filters, larger chips result in fewer edge cases. 

\subsection{CNN feature extraction}

To learn the texture representation we utilise a CNN where the learning function considers the C-Means classification \citep{bezdek1984fcm} for an objective discrimination of features. The CNN of choice was a U-Net, with 5 layers each for the encoder and decoder. Starting with 64 kernels in the encoder, these are doubled for each consecutive layer to 1024 in the final layer and opposite for the decoder. The target of the model is the fuzzy C-Means predictions. We used 2, 8 and 18 clusters to explore the performance of the U-Net.
As each fuzzy classification is non-exclusive, a multi-label loss function is used. The loss function combined Dice loss and binary cross entropy \cite {SEALE2022113044}. 

$$d(i) = 2 \times  \frac{\sum (X_i . Y_i )}{\sum(X_i) + \sum(Y_i)}$$

$$D = \frac{1}{C} \sum_{i=1}^{C}d(i)$$

As each class has its respective probability we can calculate the loss for each class, $d(i)$ and average, $D$. Here $d(i)$ is the dice loss for one class $i$, $X_i$ is the true probability and $Y_i$ is the predicted. Finally $C$ is the number of classes. Similarly, we calculate binary cross-entropy loss concerning each class and average. The final loss combining the two is a variation of combo loss \citep{taghanaki2019combo}. 

The final convolutional layer of the CNN produces the same image shape as the input with 64 convolutional activation maps in contrast to the 12 spectral channels. To combat over-fitting we applied random rotation and noise to the training set to increase the variation artificially. In addition, we applied early stopping when the loss was no longer reducing after 15 epochs. There is an increase in loss as the number of fuzzy C-means classes increases as we discriminate between even finer spectral detail.

\subsection{Graph Construction}
\label{sec:construction}
Nodes for the graphs are determined by the Simple Linear Iterative Clustering (SLIC) algorithm \citep{slic} where each segment is considered a new node. SLIC produces local homogenous parcels which for encoding geographical neighbours is ideal. 
We set the target number of segmentations for SLIC as $N = 500$. This choice was made to over-segment the image, resulting in a roughly $1310\,m^2$ coverage for each segment, which provides finer granularity while balancing computational cost. Although SLIC dynamically adjusts $N$ based on each sample as it splits or merges segments, we selected $N = 500$ as a practical compromise between segment size and efficiency. This decision was supported by sensitivity analysis (Section~\ref{sec:sensitivity}), where $N = 500$ consistently achieved the best trade-off between accuracy and performance compared to $N = 200$ and $N = 800$. We used SLIC for super-pixel generation due to its computational efficiency and ability to produce spatially compact, homogeneous segments, which are well-suited for encoding geographical neighbourhoods in graph structures. SLIC is widely used in both computer vision and remote sensing because it offers fine control over segment size and compactness while maintaining low memory overhead, an important consideration for large Sentinel-2 tiles. Although Sentinel-2 imagery contains multiple spectral channels, we applied SLIC to the RGB bands only. This choice reflects a trade-off between segmentation quality and computational cost: RGB channels capture dominant spatial and textural patterns for defining segment boundaries, while the full multi-band spectral information is incorporated later during CNN feature extraction.

The features of each node are $F_i = \{\bar{A} \}$. Where $\bar{A}$ is the mean values of each of the 64 activation maps from the CNN for each segment.
Edges for each segment are determined by connecting to the $K$ geographically nearest segments, where $K = 8$. This strategy was chosen to preserve spatial context, which is critical for remote-sensing semantics: for example, distinguishing a water segment surrounded by ocean from one adjacent to urban areas. Using feature similarity based K-NN would risk linking distant regions with similar textures, losing neighbourhood information that our pipeline relies on for context-aware embeddings. Geographical adjacency also offers computational advantages, as edge connections remain fixed and do not require iterative recomputation during training. The choice of $K = 8$ was supported by sensitivity analysis (Section~\ref{sec:sensitivity}), where this value achieved the best balance between accuracy and efficiency. This work differs from previous approaches by \citeauthor{rs14020305} \citep{rs14020305} because we only include information from the CNN and rely on geographical neighbours to construct edges. The resulting representation is a graph $G(V,E)$ with $N = V$ nodes, each with 64 features $F$. There are $K \times N$ unweighted edge connections between nodes, approximately $8 \times 500$ edges.

\subsection{Graph Neural Network}

The graph attention network is 3 layers with attention mechanisms \citep{brody2021attentive}. The final layer produces the $C$ dimensional output for each class predicted. The loss function is categorical cross entropy for each predicted class compared to the averaged C-mean of each segment. The new hidden feature vector for each node, $\mathbf{X'}_{i}$ can be calculated with the following:

\[
\mathbf{X'}_{i} = \big\Vert_{k=1}^{K} \sigma \left( \sum_{j \in{N(i) \cup{i}}} \alpha_{ij} \mathbf{W}^{k} \mathbf{x}_{j} \right)
\]
where 
\( k\) is the number of attention heads, 
\(\big\Vert\) denotes vector concatenation, 
\(\sigma (\cdot)\) is an activation function, 
\(N \) is the Neighbourhood of edges to $i$,
\(\mathbf{W}^k\) is a matrix of parameters for the \(k\)-th attention head and 
\(\alpha_{ij}\) are attention coefficients defined by the following:

\[
\alpha_{ij} = \frac{\exp \left(\mathbf{a}^T \text{LeakyReLU} \left(  \left[ \mathbf{W} \mathbf{x}_{i} \Vert \mathbf{W} \mathbf{x}_{j} \Vert \mathbf{e}_{ij} \right] \right) \right)}{\sum_{k \in N(i)\cup {i}} \exp \left( \mathbf{a}^T \text{LeakyReLU} \left(  \left[ \mathbf{W} \mathbf{x}_i \Vert \mathbf{W} \mathbf{h}_k \Vert \mathbf{e}_{ik} \right] \right) \right)}
\]

\noindent where $\mathbf{e}_{ik}$ is an edge between node $i$ and connected node $k$.

Each hidden feature vector for a node has 64 variables, given that our ultimate goal is to visualise and compare similarities we need to maintain or reduce the size of the feature vector in favour of improved computational performance over accuracy. This is achieved by reducing the output feature dimensions of the GAT layers. In our experiments, the extracted features are the output of the second to last GAT layer, within this layer we constrict the output of the features to 8 hidden feature vectors with 8 attention heads. Therefore our final representation is a 64-dimensional vector. The third layer reshapes this vector to enable comparison with the targets in the loss function.

Similarly we tested a 3 layer graph convolutional network (GCN) with an embedding vector of 60 hidden features. The choice of testing both networks is to see if the inclusion of an attention mechanism impacts the information in each node in relation to its neighbours.

For these and the above parameters we systematically explored various parameter configurations (e.g., number of GNN edges, U-Net and GCN layers, nearest neighbourhood for SLIC) and selected those that provided the best trade-off between accuracy and computational efficiency, focusing on parameters that had the most significant impact on performance (see Sections \ref{sec:sensitivity} and \ref{sec:ablation}).

\subsection{Graph Matching}
For each chip in our pipeline, we obtain a graph $G(V,E)$ where each node $V$ is linked by edge $E$. A node refers to a singular segmentation within the chip (see the yellow outlined segments in Figure~\ref{fig:pipeline draft}). Edges are connections to the nearest neighbouring nodes in feature space, $X$ from the second layer of the GNN. The second layer is chosen as it provides contextual information for each segment as it encodes neighbours from two hops away, neighbours of neighbours (this is evaluated in Section \ref{sec:contextaware}). Graph matching finds how similar two graphs are based on their node and edge composition. As our approach utilised many segments our resulting nodes are in the hundreds, therefore, utilising multi-graph solvers which take into account both vertices and edges is infeasible in both memory and computational complexity. We simplify the matching to a bi-partite graph representation using the Hungarian algorithm to match nodes present in both graphs and calculate the overall similarity based on the distance between matched nodes in feature space $X$. Applying the matching to create a similarity matrix for every chip processed through UMAP \citep{mcinnes2020umapuniformmanifoldapproximation} creates a two-dimensional embedding for visualisation as follows in Section \ref{sec:projection}.

\subsection{2D Projections}
\label{sec:projection}
We provide the user an interface that can switch between two modes. In the higher-level mode, users interact with $256 \times 256$ pixel chips. In the lower-level mode, users interact with the segments. In both cases, the user interacts through a 2D UMAP projection of the manifold -- Figure \ref{fig:overview} (top) -- and sees the relevant chips or segments in the view below -- Figure \ref{fig:overview} (bottom).

UMAP requires either a set of vectors representing each data point or a precomputed similarity matrix, typically based on pairwise distances or cosine similarities. Low-level segments are represented by a singular 64 or 60-dimensional vector from the graph attention networks and are readily available for UMAP projection. Higher-level chips are represented as graph structures, where each node corresponds to a 64 or 60-dimensional vector. To compare two chips, we apply Hungarian matching to align their nodes, producing a one-to-one correspondence between vectors. We then compute the cosine similarity for each matched pair and average these values to obtain an overall similarity score between the two chips. Repeating this process across all chip pairs yields a similarity matrix, which serves as input to UMAP for generating the high-level 2D projection.

\paragraph{Front End (User Interface).}
The front end provides the interactive 2D manifold view (UMAP) at two levels: (i) chip-level points, and (ii) segment-level points for selected images. Users can pan, zoom, and brush rectangular selections. Selected points are reflected in a linked image panel that renders the corresponding chip thumbnails or super-pixel masks. For performance, the user interface is written using \texttt{DirectX 11} and \texttt{C++}. The real-time interaction (in the accompanying video) was captured whilst using an NVIDIA RTX 3090.

\paragraph{Back End (Model and Data Handling).}
All model training and inference are implemented in \texttt{PyTorch} with GPU acceleration via \texttt{CUDA}. The offline pipeline includes U-Net training for feature extraction, GNN training for context-aware embeddings, and UMAP projection for dimensionality reduction. These activation maps, graph embeddings, and 2D coordinates, are saved to files and accessed by the front-end interface for interactive exploration. During interaction, the tool reads these precomputed files to render chip thumbnails, segmentation masks, and manifold positions in real time. This design separates heavy batch computation (performed offline on an NVIDIA RTX 3090 GPU) from lightweight rendering and selection operations.

\subsection{Evaluation}
To test the GNN part of our architecture, we define two levels of similarity evaluation between the SLIC super-pixels (segments) -- feature-based and context-aware.

\subsubsection{Graph Evaluation I: Feature-based}
\label{sec:graphevaluation}
Under feature-based evaluation the SLIC segments are fed into each GNN, producing feature space, $X$. For each segment, we compare it to its closest segment within the feature space. The comparison was made using four common similarity measures; gray level occurance matrix (GLCM), linear binary patterns (LBP), Structured similarity index measure (SSIM) and spectral angle mapper (SAM). The first two measures were calculated via a bounding box that enclosed the segment, the latter two purely on the content within the segment. This test allows us to find the conditions and parameters that produce the best GNN.

\subsubsection{Graph Evaluation II: Context-aware}
\label{sec:contextaware}

Under context-aware evaluation, we enhance the evaluation by considering spatial context. For each segment \(x\), we evaluate its similarity not only to \(y\) in feature space but also by taking into account the 8 nearest spatial neighbours of both segments. The similarity between \(x\) and \(y\) is then adjusted to ensure that the spatial arrangement of their neighbours is also preserved. We use the Hungarian matching algorithm to find the optimal one-to-one correspondence between the neighbours of \(x\) and those of \(y\), to test neighbourhood preservation. This context-aware evaluation leads to a more accurate representation of the similarity between super-pixels, reflecting both feature similarity and spatial proximity. The one-to-one correspondence resulting from the bi-partite matching, leads to 9 nodes pairs, each pair consisting of two segments, either \(x\) or a neighbour of \(x\) and \(y\) or a neighbour of \(y\). Each pair's similarity is then computed with GLCM, LBP, SSIM and SAM and averaged for all nine pairs.

\subsubsection{U-Net Evaluation}
\label{sec:unetevaluation}
Part of our architecture is a U-Net that provides unsupervised features to feed into the GNN and graph matching. For this evaluation we compare our U-Net architecture to the state-of-the-art approaches of using ResNet-50 and GoogLeNet \citep{helber2019eurosat}. The benchmarked approaches \citep{helber2019eurosat} were pre-trained on millions of images and then fine-tuned on EuroSat data. We utilise our pre-trained U-Net with two additional linear layers appended for classification. Our U-Net being pre-trained on only $1411$ images, from a singular geographical region. If this test achieves an accuracy close to state-of-the-art we can attest that our U-Net, with some fine-tuning, has learnt rich features for the remaining parts of our pipeline. Successful application of the U-Net on a larger and more geographical diverse dataset also would demonstrate geo-generalisable weights.

\subsubsection{Tool Evaluation}
\label{sec:toolevaluation}
Additional to the above approaches, we also provide an evaluation of our interactive application that allows the user to explore and label the final features produced in two dimensions via UMAP for exploration and labelling. \textbf{Please refer to the video submitted as a supplementary file} for an example exploration of the feature space to label at both the top chip level and the segmentation level. A public version is available at 
\blindtext{\href{https://youtu.be/GZl1ebZJgEA}{YouTube}, and is archived at \href{https://doi.org/10.5281/zenodo.16676591}{Zenodo}~\cite{tulsi2025labellingvideo}.}
{ \href{Hidden}{YouTube}, and is archived at \href{Hidden}{Zenodo}.}

\section{Results}
\subsection{Remote Sensing Labelling Application (see \ref{sec:toolevaluation} Tool Evaluation)}

\vspace{-3mm}
\begin{figure}[ht]
  \centering
   \includegraphics[width= 0.6 \textwidth ]{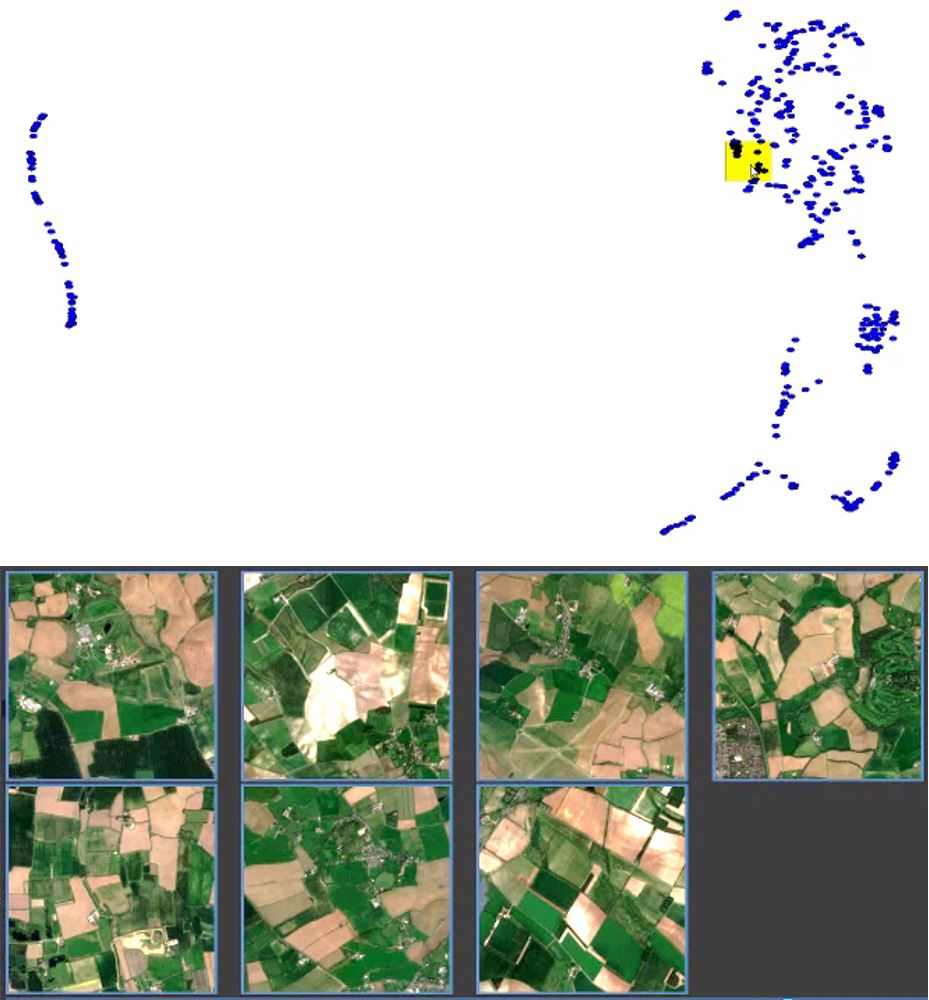}
   \caption{UMAP dimension reduction to 2D on the graph matching output of our entire pipeline. At this level, each point represents one chip. The user interactively highlights a resizeable region which can be dragged across the manifold representation. Chip images represented by the 2D points within the highlight are displayed in the pane below. Training did not use any labelled data. (Video frame edited to save space in the paper).}
   \label{fig:overview}
\end{figure}

Our pipeline enables advanced interaction within our labelling tool. Referring to the \textbf{provided video} \blindtext{\cite{tulsi2025labellingvideo}}{}
of the tool utilised on the dataset, this section presents frames extracts from the video to demonstrate and discuss functionality.

\begin{figure*}[ht]
  \centering
   \includegraphics[width= 0.95 \textwidth ]{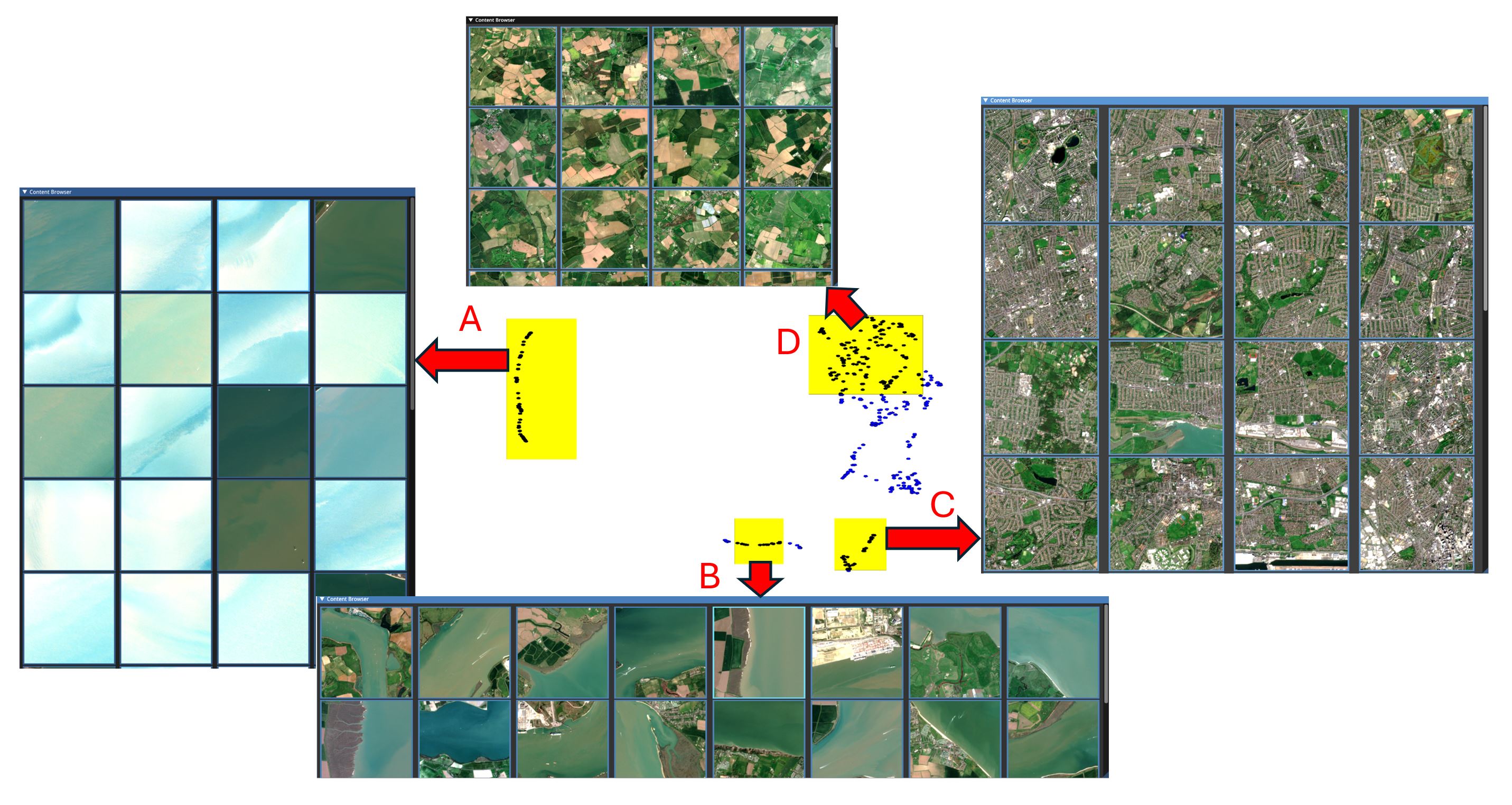}
   \caption{Example of the U-map embedding dimensionality reduction of high dimensional feature space, $X$, output from the final graph matching stage of our entire pipeline. Training was unsupervised. The images selected by the user highlights are shown in the display pane with the labels A-D explained in Section \ref{sec:clusterexplore}.}
   \label{fig:8Annotated}
\end{figure*}

The main contributions are \textbf{(Section \ref{sec:clusterexplore})} the tool allows exploration of feature space, $X$, via a two-dimensional interface where our pipeline has resulted in successful organisation of the high dimensional manifold. The resultant clusters are compact and contain visually related images. \textbf{(Section \ref{sec:rotation})} embedded images with large texture gradients have no adverse impacts on the similarity measure using our approach. \textbf{(Section \ref{sec:segmentationanalysis})} we significantly extend previous labelling approaches by enabling this interaction on pixel-based segmentations within the larger images, thus allowing fast labelling at a finer granularity.

\subsubsection{Cluster exploration}
\label{sec:clusterexplore}

In Figure \ref{fig:overview} (top), the user interacts with the 2D manifold representation of the $256 \times 256$ chip-level data by brushing selected points. The brushed points, highlighted in yellow, correspond to specific source images, which are simultaneously displayed in the image pane (Figure \ref{fig:overview}, bottom). This particular cluster reveals a patchwork of fields. The accompanying video illustrates the manifold's spatial evolution by showing image chips transitioning through farmland, industrial zones, residential areas, increasing water bodies (e.g., lakes and sea), and other geographical features.

Figure \ref{fig:8Annotated} shows more example content of the embedding space. Samples drawn and shown from the clusters contain mainly sea (A), land with water (B), dense built-up areas (C) and farmland (D). Area A in the embedding is consistently majority water content, there is very little presence of land. However, the features within that cluster do not differentiate between any features such as boats or offshore installations. Sample space B includes nearly all water content features present near land, the cluster sits directly between A and D which are exclusively just land or water. This evolution of features shows at the chip level that water content is a highly relevant feature for similarity. 

\begin{figure*}[ht]
  \centering
   \includegraphics[width= 0.85 \textwidth ]{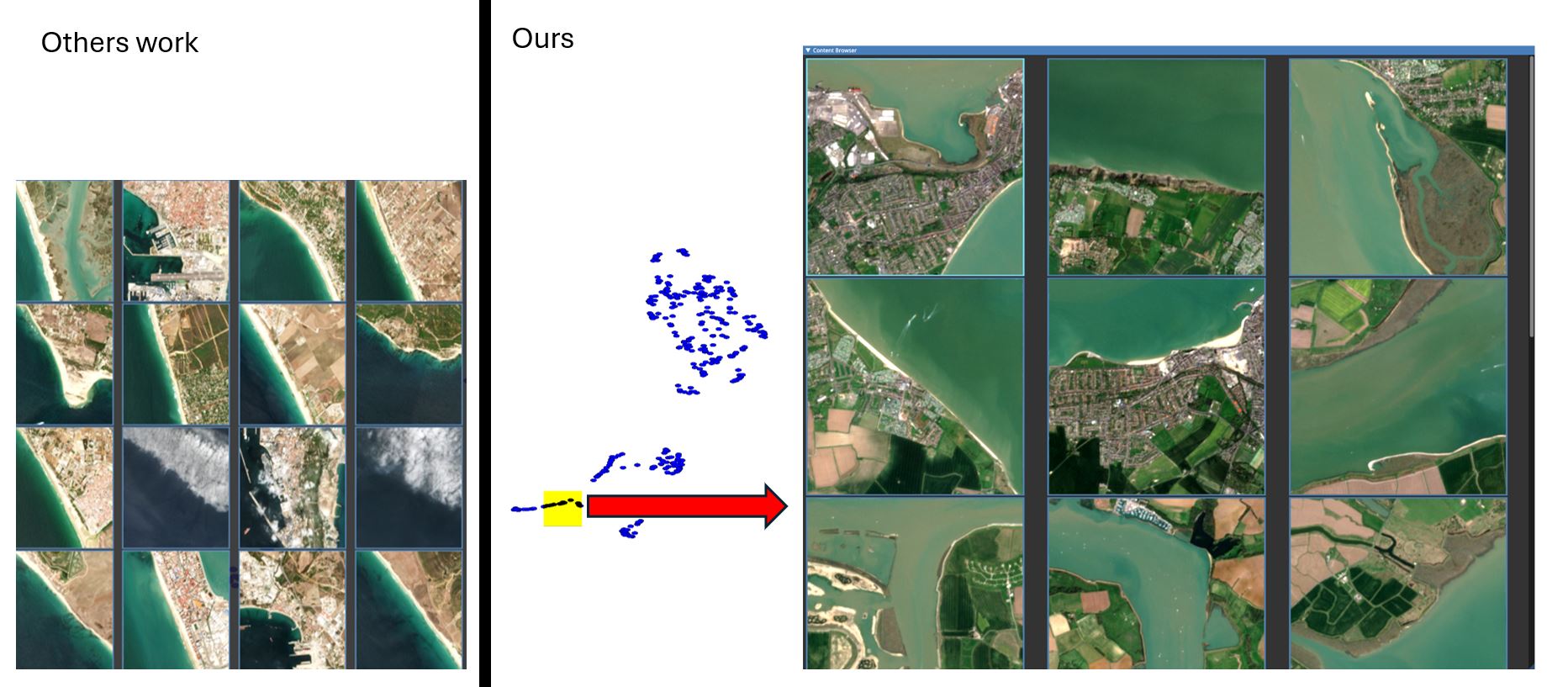}
   \caption{Example of the UMAP embedding space from \protect\citep{patel2023manifold}, showing results influenced by strong texture similarities due to edge alignment, and a comparison with our approach, which introduces rotational invariance to remove this strong alignment.}
   \label{fig:rotation}
\end{figure*}

\subsubsection{Rotational Invariance (Figure \ref{fig:rotation})}
\label{sec:rotation}
Within our application, the utility of representing our images as segmentations with mean feature aggregation has a two-fold effect. Firstly, rotational invariance is introduced by Hungarian matching disregarding the geographical layout of features, as only segments are compared. Secondly, taking the mean of each feature map removes any encoding of strong gradient changes within the activation maps. For an example of both, we refer to Figure \ref{fig:rotation}. Work produced by \citep{patel2023manifold} shows an example of a large textural and spectral disparity between land and sea. The clustering has been directly influenced by the orientation of this boundary, with the inclusion of outliers containing cloud and water where they follow the same gradient. In comparison our work has ignored strong directional gradients within the image data, e.g. see Figure \ref{fig:8Annotated}. In Figure \ref{fig:rotation}, we visually selected a cluster also of approximately 50/50 land/water coverage, which indicates there is no obvious common directional gradient.

\begin{figure}[ht]
  \centering
   \includegraphics[width=0.8\textwidth ]{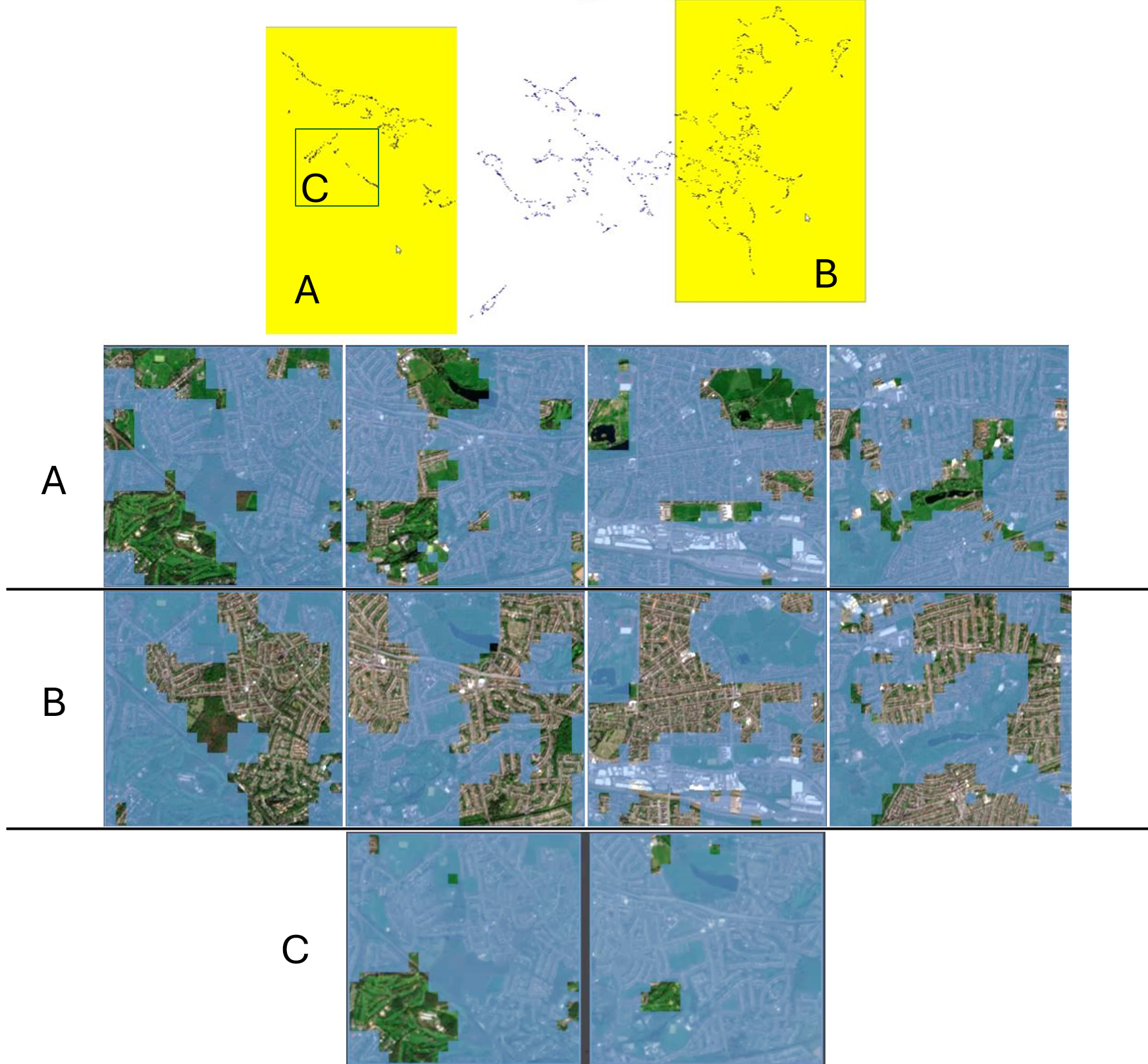}
   \caption{Example (from video) of projecting four images and exploring their segmentations to label as urban or vegetation. The left selection, (A), shows the largely vegetation segments in the first row of images and (B) largely urban development in the same four images, but demonstrated in the second row. (C) is a selection in the manifold of segmentations which related to golf courses and are present on different images as displayed on the third row (the golf course example is in the video).}
   \label{fig:urban}
\end{figure}

\subsubsection{Segmentation Analysis (Figure \ref{fig:urban})}
\label{sec:segmentationanalysis}

A major contribution is enabling finer sub-image labelling based on the segmentations. The user starts by selecting images they wish to examine in detail. A new interactive 2D projection of the feature space, $X$, is displayed where each point represents a single (SLIC) segmentation within the images. Similar to the image exploration, our approach produces a successful organisation of the manifold where segments are very closely related. In Figure \ref{fig:urban} we see at the top the 2D interactive display where each point represents one segment within the image data. By brushing multiple points in 2D, the user will see all corresponding segmentations in the reference images below. This manifold evolves from vegetation on the left to largely urban areas on the right. Here we combine two frames from the video where the left selection displays the vegetation with some urban features (top row of images), and the right selection displays the almost pure urban features (bottom row of images). This empowers the expert to quickly label at a fine level of segmentation (as demonstrated in the video).

The third row shows a further example of extracting features that share common geographical neighbourhoods, in this particular example golf courses. Although unsupervised, the embedding space has clustered larger and smaller golf courses irrespective of their sizes. The video demonstrates an example of how to build up a labelling of the golf course data through selection and filtering data points in the manifold.

\begin{figure}[ht]
  \centering
   \includegraphics[width= .8 \textwidth ]{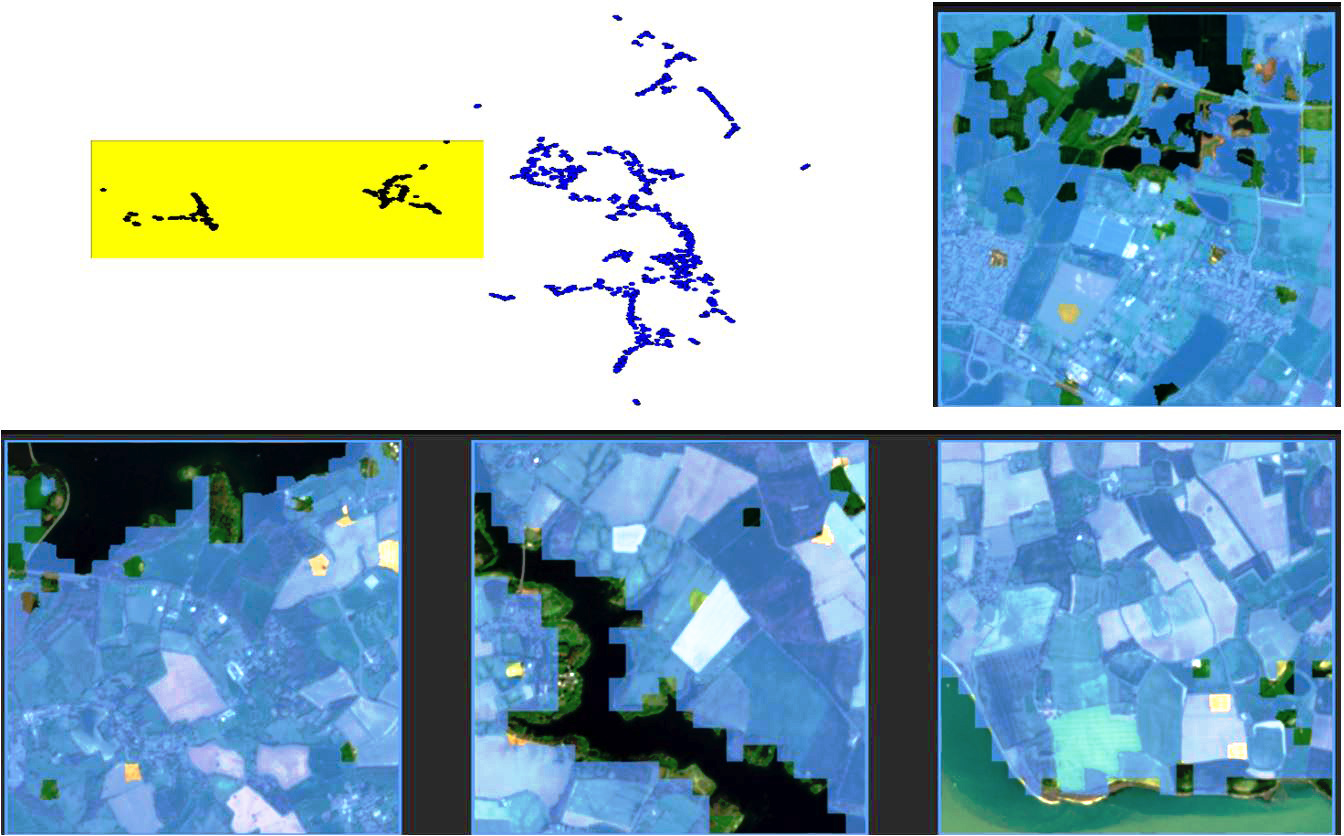}
   \caption{Example of extracting water features at segment level. There are various water features including lakes, rivers and sea. Here, as in Figure \ref{fig:urban}, the semi-transparent blue indicates the areas masked away, and the full colour segments (which are dark because they are water features) are the selected areas for labelling.}
   \label{fig:water}
\end{figure}

The projection at segment level has also clustered the various water bodies within the images (figure \ref{fig:water}). Via exploring the manifold, the user is able to make a selection that segments water features effectively. We found that various different forms and sizes of water bodies including large bodies of water are easily selected via the interface and extracted simultaneously.

\subsection{U-Net Test (proposed in U-Net Evaluation \ref{sec:unetevaluation})}

The aim of this test is to evaluate whether the U-Net part of our architecture achieves close to state-of-the-art performance to feed into the rest of our graph neural networks, graph matching and interactive application pipeline.

\begin{table}[ht]
  \centering
  \caption{Classification accuracies for each model tested, where $C$ is the initial Fuzzy C-Means clustering number. EuroSat results are taken from \citep{helber2019eurosat}. These results serve as a validation of the U-Net component rather than the primary contribution, confirming its ability to learn generalisable features for use in our unsupervised labelling pipeline.}  \begin{tabular}{@{}lcc}
    \toprule
    Method & Accuracy  \\
    \midrule
    EuroSat ResNet-50           & 98.57 \\
    EuroSat GoogleLeNet         & 98.18 \\
    C=2 linear layers           & 93.24 \\
    C=8 linear layers           & 97.43 \\
    C=18 linear layers          & 93.42 \\
    \bottomrule
  \end{tabular}
  \label{tab:resCNN}
\end{table}

In order to validate the utility and generalisable application of the methodology we want to demonstrate that the trained U-Net (trained using unsupervised C-means clustering) is generalisable. In order to evaluate the U-Net we compare to a benchmark experiment on the EuroSat dataset \citep{helber2019eurosat}. The dataset consists of 27,000 labelled images from Sentinel-2 mission covering ten separate classes. We resized the images from 64x64 to 256x256 to match the training data. We did not apply bottom-of-atmosphere correction. The Eurosat models were pre-trained on 1.2 million images from the ILSVRC-2012 dataset.

We append and train two additional linear layers for class prediction on the EuroSAT dataset. Our aim is to evaluate if the learnt weights are generalisable and whether the model can find an appropriate minimum given a starting feature space. Despite not using a traditional classification model (we use a U-Net) and pre-training on a much smaller dataset, our approach achieves performance comparable to state-of-the-art methods (Table \ref{tab:resCNN}). This validates the U-Net component of our architecture.

It is important to note that Table~\ref{tab:resCNN} is not intended as a primary result of this work but as a validation step to demonstrate that our U-Net, trained in an unsupervised manner, learns generalisable features suitable for integration into the full pipeline.

\subsection{Graph Encoding Comparison (see \ref{sec:graphevaluation} Graph Evaluation)}

We trained GCN and GAT variations of graph networks against C-means clusters with C=2, 8 and 18. These were evaluated using various similarity measures. GLCM and LBP indicate the textural differences between images. SSIM measures structure, contrast and luminance. SAM is based solely on the mean spectral difference between each segment. This section analyses the construction of our embedding spaces by determining the effect of the different C-means cluster number on the pipeline. 

\begin{table}[ht]
  \centering
  \caption{Average of each similarity measure over every pair of nearest neighbouring segments in feature space, $X$, created by each method. This corresponds to Section \ref{sec:graphevaluation} Feature-based graph evaluation.}
  \begin{tabular}{@{}lcccc}
    \toprule
    Model & GLCM$\downarrow$ & LBP$\uparrow$ & SSIM$\uparrow$ & SAM$\downarrow$ \\
    \midrule
    GCN 2   & 16.2770           & 0.8571            & 0.9120            & 0.3321 \\
    GCN 8   & \textbf{15.4685}  & \textbf{0.8716}   & \textbf{0.9287}   & \textbf{0.2887} \\
    GCN 18  & 15.5322           & 0.8671            & 0.9243            & 0.3017 \\
    GAT 2   & 16.0025           & 0.8633            & 0.9201            & 0.3112 \\
    GAT 8   & 15.5676           & 0.8704            & 0.9256            & 0.2942 \\
    GAT 18  & 15.7663           & 0.8673            & 0.9237            & 0.2992 \\
    \bottomrule
  \end{tabular}
  \label{tab:resGraphS}
\end{table}

\begin{table}[ht]
  \centering
  \caption{Average of each similarity measure over every pair of nearest neighbouring segments. The similarity between each segment takes into account the context of its local geographical neighbourhood as described in the referring text. This corresponds to Section \ref{sec:contextaware} Context-aware graph evaluation.}
  \begin{tabular}{@{}lcccc}
    \toprule
    Model & GLCM$\downarrow$ & LBP$\uparrow$ & SSIM$\uparrow$ & SAM$\downarrow$  \\
    \midrule
    GCN 2   & 14.4610           & 0.8983            & 0.9589            & 0.2152  \\ 
    GCN 8   & \textbf{13.7904}  & \textbf{0.9034}   & \textbf{0.9624}   & \textbf{0.2064}  \\
    GCN 18  & 13.7278           & 0.9033            & 0.9620            & 0.2066  \\
    GAT 2   & 14.2259           & 0.9000            & 0.9560            & 0.2122 \\
    GAT 8   & 14.0145           & 0.9018            & 0.9613            & 0.2092 \\ 
    GAT 18  & 14.0557           & 0.9013            & 0.9611            & 0.2100 \\
    \bottomrule
  \end{tabular}
  \label{tab:resGraphN}
\end{table}

This first experiment computes the measures between each segment, A, and its nearest neighbour, B, in the feature space $X$ (Table \ref{tab:resGraphS}) and average that over all the segments. This tests whether the GNN places similar segments together in the feature space. The number of C-means clusters exhibits low sensitivity, and we therefore fix C=8 as the optimal choice for our model.

The second experiment will find the eight spatially closest segmented neighbours in the original image to A. It also finds the eight spatially closest segment neighbours to B. Hungarian matching is then used to pair segments between these two neighbourhoods, forming nine matched pairs. The similarity measures are computed for each pair to assess how similar the neighbourhoods are. This experiment tests whether segments with similar local contexts, implicitly captured by the two graph layers, are also close in the learned feature space. Ideally, the feature space should position segments near others that share similar neighbourhood structures. For example, a small body of water surrounded by urban segments should be placed near other similarly situated water bodies.

Overall the aim of this test is to evaluate that our pipeline is able to place segments of a certain type and surrounding close within the feature space as this lends itself to being the best interaction for a user by enforcing local structural awareness.
The best similarity results were for eight clusters and GCN type was better than GAT, see Table \ref{tab:resGraphN}. From our results presented in the video it is possible to see that segments are contextually close making the labelling easier.
The graph convolutional network, trained on eight C-means clusters seems to be the best balance in our tests and was therefore used in our pipeline, although again there is low sensitivity to cluster number. Overall these models take into account the neighbourhood and texture more than the individual segments or spectral differences.

\subsection{Parameter Sensitivity}
\label{sec:sensitivity}
In this section we explore the parameter sensitivity for two key values within the pipeline. These two parameters are the initial $K$ value and $N$ for SLIC during the graph construction phase, Section \ref{sec:construction}. The variations in parameters are tested via our context-aware approach to segmentation comparison detailed in Section \ref{sec:contextaware}. We also report results when utilising only the CNN output and the first layer of the GCN. These tests are compared against the best pipeline (GCN 8) where $K=8$ and $N=500$ segments in the SLIC algorithm.
\begin{table}[ht]
  \centering
  \caption{Comparison of varying $K$, during the graph construction phase (Section \ref{sec:construction}).}
  \begin{tabular}{@{}lcccc}
    \toprule
    K & GLCM$\downarrow$ & LBP$\uparrow$ & SSIM$\uparrow$ & SAM$\downarrow$  \\
    \midrule
    4    & 14.2846           & 0.8992            & 0.9583            & 0.2140  \\ 
    8    & 13.7904  & 0.9034   & 0.9624   & 0.2064  \\
    12   & \textbf{13.2240}   & \textbf{0.9075}    & \textbf{0.9649}   & \textbf{0.1988}   \\
    \bottomrule
  \end{tabular}
  \label{tab:ablationK}
\end{table}

We used 8 edges during the graph generation phase. Here we compare results for $K\in\{4,8,12\}$. Our pipeline shows an improvement in for larger number of edges ($K$) in the neighbourhood (Table \ref{tab:ablationK}). 
Increasing the number of edges ($K$) increases the GCN training time and graph matching. For example, given the Hungarian algorithm's complexity of $O(N^3)$, where $N$ is the sum of all nodes in both graphs, the complexity for $K=8$ is $16^3$ whilst $K=12$ is $24^3$. Correspondingly, for the GCN training times for $K=8$ was $1.6$ hours and $K=12$ was $6.6$ hours. In our tests, larger $K$ produces more accuracy, but at significantly slower training times, and therefore we used $K=8$ as a good balance.

\begin{table}[ht]
  \centering
  \caption{Comparison of varying SLIC parameter $N$, used during the graph construction phase Section \ref{sec:construction}.}
  \begin{tabular}{@{}lcccc}
    \toprule
    SLIC & GLCM$\downarrow$ & LBP$\uparrow$ & SSIM$\uparrow$ & SAM$\downarrow$  \\
    \midrule
    200     & \textbf{12.4336}           & 0.8888            & 0.9166            & 0.3608  \\
    500    & 13.7904  & \textbf{0.9034}   & \textbf{0.9624}   & \textbf{0.2064}  \\
    800     & 14.8003           & 0.8604            & 0.9404            & 0.2819 \\
    \bottomrule
  \end{tabular}
  \label{tab:ablationSLIC}
\end{table}

Here we report values of $N\in{200,500,800}$ for the number of SLIC segments. Our initial choice, $N=500$, was to cover an approximate area of $1300m^2$ for ease of interpretability and efficiency when visualised for the user (the mean average size of segments is $256 \times 256 \div 500 = 131$ --- approximately 11 by 11 pixels and each pixel is $10m^2$ in Sentinel-2). A smaller area allows finer labelling at the cost of speed, namely projecting and labelling more points. This choice for $N$ was also optimal when considering local neighbourhood encoding using the different measures, see Table \ref{tab:ablationSLIC}. For smaller $N$ values, we see that comparing segments via GLCM shows better results however, over all measures $N=500$ is the most accurate.

\subsection{Graph layer ablation}
\label{sec:ablation}

\begin{table}[ht]
  \centering
  \caption{Comparison of utilising different outputs for the final graph representation, the graph construction phase Section \ref{sec:construction} and the first layer of the GCN. These are compared to the ideal pipeline, GCN with 8 C-means.}
  \begin{tabular}{@{}lcccc}
    \toprule
    Image Representation& GLCM$\downarrow$ & LBP$\uparrow$ & SSIM$\uparrow$ & SAM$\downarrow$  \\
    \midrule
    Graph Generation  & 14.6613           & 0.8970            & \textbf{0.9853}   & 0.2165  \\
    Layer 1           & 14.3951           & 0.8988            & 0.9590            & 0.2147 \\
    GCN Layer 2       & \textbf{13.7904}  & \textbf{0.9034}   & 0.9624            & \textbf{0.2064}  \\
    
    \bottomrule
  \end{tabular}
  \label{tab:ablationOutput}
\end{table}

Finally, we validate our architecture by conducting an ablation study on the use of embeddings from our graph output.
Our architecture produces graph outputs at three stages: (1) during graph generation, which combines the CNN U-Net output and SLIC; (2) at the output of GCN layer 1; and (3) at the output of GCN layer 2, which is the version reported throughout this work. The output of GCN layer 3 is a learnt reshaping from the second layer, from hidden dimension to C-means predictions, used only for the loss function. As such, it is not suitable for use as an embedding.

Table \ref{tab:ablationOutput} demonstrates that selecting GCN layer 2 as the embedding layer yields optimal performance on three of the metrics related to texture and spectral similarity, which are particularly relevant to our interactive pipeline. The only exception is SSIM, where the combination of CNN U-Net output and SLIC performs best.

\subsection{Computational Complexity and Runtime Efficiency}
The proposed pipeline was designed with efficiency in mind, balancing accuracy and computational cost through parameter tuning (Section~\ref{sec:sensitivity}). The complexity of each stage is as follows: SLIC segmentation operates in approximately $O(P)$ time, where $P$ is the number of pixels, and is highly efficient for large Sentinel-2 tiles. Fuzzy C-Means clustering introduces iterative updates but remains tractable for moderate cluster counts ($C \in \{2,8,18\}$). Using THOP, our U-Net requires 55 GFLOPs per forward pass for a 256×256 input with 12 Sentinel-2 channels. The model contains 31M trainable parameters. Graph construction scales with $O(N + K \cdot N)$, where $N$ is the number of super-pixels and $K$ the number of edges per node (here $K = 8$). Message passing in the GNN is linear in the number of edges, $O(|E|)$, and was implemented with efficient batching. The GNN contains 17K parameters and requires 415 MFLOPs per forward pass. Hungarian matching introduces the highest theoretical complexity at $O(n^3)$ for $n$ nodes per graph (approximately $n \approx 500$), but this step is applied only during similarity computation and is GPU-accelerated. Dimensionality reduction via UMAP scales well for interactive use and was precomputed for large batches.

On our test hardware (NVIDIA RTX 3090 GPU, 24 GB RAM), training the U-Net required approximately 10 hours for 600 epochs, and the GNN training completed in 1.5 hours for 300 epochs. Inference for a single tile ($42\times 42$ chips comprised of $256 \times 256$ pixels) through the pipeline (segmentation, feature extraction, graph generation, graph inference, Hungarian matching, and embedding) requires approximately 23 minutes. This time is dominated by the $O(n^3)$ Hungarian matching process (194,481 pairs of chips) which takes 20 minutes (Table \ref{tab:timings}), however is only computed once for a dataset. The application interaction achieves a frame rate update of 60 frames per second, although this is variably dependent on the amount of points selected by the user via the brushing interaction.

\begin{table}[h!]
\centering
\caption{Example runtime breakdown of the full processing pipeline on $1764$ chips (each $256 \times 256$ pixels).}
\begin{tabular}{l r}
\hline
\textbf{Step} & \textbf{Time} \\
\hline
Segmentation & 2 min 20 s \\
Feature extraction (U-Net) & 6 s \\
Graph generation (parallelised) & 8 s \\
GNN & 6 s \\
Hungarian matching, (194{,}481 pairs, parallelised) & 20 min \\
Embedding (UMAP) & 15 s \\
\hline
\end{tabular}
\label{tab:timings}
\end{table}

\section{Discussion}

Our pipeline creates a 2D interactive interface that enables users to apply labels to segments and chips extracted from remote sensing (RS) images. We demonstrate that unlabelled data can be effectively leveraged using C-means clustering, SLIC segmentation, and a U-Net architecture, all of which contribute to the loss function of a graph convolutional network (GCN). The GCN takes as input the SLIC segments and the final layer of the U-Net, allowing it to learn representations that capture both individual segment features and their spatial context. The resulting graph embeddings are compared using Hungarian matching to assess neighbourhood similarity. These comparisons are used to construct a similarity matrix, which is then projected into 2D using UMAP to power the interactive interface.

By exploring hyperparameters, we arrive at an optimal model and architecture that, as shown in the accompanying video, enables intuitive and efficient interaction with both high-level chips and low-level segments. We validated key components of the model and provided a quantitative assessment of its performance including sensitivity and ablation study.

The high ingestion rate of remote sensing (RS) data and the need for labelled datasets remain major challenges -- especially in emerging or niche application areas where accurate, up-to-date labels are scarce. Creating large, high-quality datasets typically requires costly expert input. To address this, our proposed application framework reduces labelling effort by enabling users to find similar features across RS images. Presenting data as image chips, rather than traditional map views, streamlines the labelling of large areas. While map-based projections offer geographic context, chips allow users to more easily assess visual and contextual similarities. Future work could integrate both views through a coordinated interface.

Remote sensing (RS) data captures the physical properties of surface materials, which reflect light at varying intensities depending on the recorded wavelength. Within a single $10m^2$ pixel, multiple materials may contribute to the signal, resulting in mixed spectral responses, therefore, spectral unmixing was the original inspiration behind our generalisable CNN model. Our use of fuzzy clustering emulates this principle by allowing soft associations between segments and composite material types. However, an open question remains: could the pipeline be further improved by integrating spectral unmixing algorithms \cite{9775570} commonly used in the RS community? Alternatively, incorporating a priori knowledge of material reflectance properties could guide centroid selection to better target or generalise the labelling process.

The graph matching process within this pipeline and the tests poses an interesting computational problem. Whilst Hungarian matching can be processed on graphics cards reducing time taken it still remains a costly problem to solve. The complexity is $O(n^3)$ where $n$ is the count of nodes in both graphs. In order to save time this process could be estimated via an AI model. Similarly UMAP and t-SNE could be solved by training parametric models \cite{sainburg2021parametric,svantesson2023get}. Large dataset projections could be calculated offline to user interaction however branched projections of many segments could benefit greatly from parametric models.

At the low-level we allow users to label segments. Even lower pixel-level classification would require future work to improve the segmentations where only extremely similar or identical pixels are grouped. This could impact the time complexity of the graph matching as more segmentations would be required also placing more demand on dimensionality reduction. Alternatively segments could be organised hierarchically and presented to a user for accurate pixel level classification, therefore only computing relevant information when needed.

\section*{Acknowledgements}
This research was supported by \blindtext{EPSRC}{XXX} grant number \blindtext{EP/S021892/1}{XXX} and \blindtext{UKHO}{XXX}. For the purpose of open access the authors have applied a Creative Commons Attribution (CC BY) license to any Author Accepted Manuscript version arising from this submission.

\section*{Declaration of Interest Statement}
The authors declare that they have no known competing financial interests or personal relationships that could have appeared to influence the work reported in this paper.

\section*{Data Statement}
Data supporting this study are included within the article and its supporting materials. The work also utilised two publicly available datasets: the Eurosat dataset by \citeauthor{helber2019eurosat} \citep{helber2019eurosat} and the Sentinel-2 Water Edges Dataset (SWED) \citep{SEALE2022113044}. The Eurosat data is available publicly here (\url{https://github.com/phelber/eurosat}). The SWED dataset is available from \url{https://openmldata.ukho.gov.uk/}

\bibliographystyle{abbrvnat}
\bibliography{template}

\end{document}